\documentclass[letterpaper, 10 pt, conference]{ieeeconf}
\IEEEoverridecommandlockouts    
\usepackage{graphics}
\usepackage{times}
\usepackage{amsmath}
\usepackage{amssymb}
\usepackage{graphicx}
\usepackage{algorithm}
\usepackage[noend]{algpseudocode}
\usepackage{booktabs,makecell}
\usepackage{color}
\usepackage{listings}
\usepackage{mathtools}
\usepackage{subfiles}
\usepackage{hyperref}
\usepackage[acronyms, shortcuts]{glossaries}
\usepackage{multirow,multicol}
\usepackage[nocompress]{cite} 
\definecolor{instructioncolor}{rgb}{.5,.5,.5}

\usepackage[font=small]{caption}

\def\secref#1{Sec.~\ref{#1}}
\def\figref#1{Fig.~\ref{#1}}
\def\tabref#1{Tab.~\ref{#1}}
\def\eqref#1{Eq.~(\ref{#1})}

\makeatletter
\usepackage{xspace}
\DeclareRobustCommand\onedot{\futurelet\@let@token\@onedot}
\def\@onedot{\ifx\@let@token.\else.\null\fi\xspace}

\def\etal{{et al}\onedot}
\makeatother

\def\etalcite#1{\etal~\cite{#1}}

\usepackage{array}
\newcolumntype{L}[1]{>{\raggedright\let\newline\\\arraybackslash\hspace{0pt}}m{#1}}
\newcolumntype{C}[1]{>{\centering\let\newline\\\arraybackslash\hspace{0pt}}m{#1}}
\newcolumntype{R}[1]{>{\raggedleft\let\newline\\\arraybackslash\hspace{0pt}}m{#1}}

\def\argmin{\mathop{\rm argmin}}

\newcommand{\RR}{\mathbb{R}}

\renewcommand{\b}[1]{\mbox{\boldmath$#1$}}

\renewcommand{\d}[1]{\b {#1}}

\renewcommand{\v}[1]{{\b #1}} 

\newcommand{\mq}[1]{{\mbox{{\sffamily{#1}}}}}

\graphicspath{{pics/}}

\newcommand{\slam}{KISS-SLAM\xspace}
\newcommand{\kiss}{``Keep It Small and Simple"\xspace}
\newcommand{\helipr}{HeLiPR\xspace}
\newcommand{\pose}{\mq{T}}
\newcommand{\se}{\mathbb{SE}}

\newcommand{\pert}{\Delta \boldsymbol\omega}
\newcommand{\pred}{\hat{\mq{T}}}
\newcommand{\set}[1]{\mathcal{#1}}
\newcommand{\spoint}{\d{s}}
\newcommand{\tpoint}{\d{q}}

\newcommand{\keypose}{\pose\xspace}
\newcommand{\localmap}{\mathcal{M}\xspace}
\newcommand{\trajectory}{\mathcal{T}\xspace}
\newcommand{\cv}{\Delta \pose_t}
\newcommand{\closure}{\pose_{i\rightarrow j}}
\newcommand{\closurehat}{\hat\pose_{i\rightarrow j}}
\newcommand{\grid}{\mathcal{V}}

\newacronym{icp}{ICP}{iterative closest point}
\newacronym{slam}{SLAM}{simultaneous localization and mapping}
\newacronym{loam}{LOAM}{lidar odometry and mapping}
\newacronym{suma}{SuMa}{Surfel-based Mapping}
\newacronym{cticp}{CT-ICP}{continuous time ICP}
\newacronym{imu}{IMU}{inertial measurement unit}
\newacronym{gps}{GPS}{global positioning system}
\newacronym{ugv}{UGV}{unmanned ground vehicle}
\newacronym{mcl}{MCL}{Monte-Carlo localization}
\newacronym{srrg}{RVP}{Sapienza Robots Vision and Perception group}
\newacronym{ate}{ATE}{absolute trajectory error}
\DeclareRobustCommand{\rchi}{{\mathpalette\irchi\relax}}
\newcommand{\irchi}[2]{\raisebox{\depth}{$#1\chi$}} 

\usepackage{flushend}

\title{\LARGE \bf KISS-SLAM: A Simple, Robust, and Accurate 3D LiDAR\\ SLAM System With Enhanced Generalization Capabilities}

\author{
  Tiziano Guadagnino* \and Benedikt Mersch* \and Saurabh Gupta \and Ignacio Vizzo \and Giorgio Grisetti \and Cyrill Stachniss
}

\begin{document}

\thispagestyle{empty}
\pagestyle{empty}
\twocolumn[{
			\renewcommand\twocolumn[1][]{#1}
			\maketitle
			\begin{center}
				\fontsize{12}{12}\selectfont
				\def\svgwidth{0.99\linewidth}
\begingroup%
  \makeatletter%
  \providecommand\color[2][]{%
    \errmessage{(Inkscape) Color is used for the text in Inkscape, but the package 'color.sty' is not loaded}%
    \renewcommand\color[2][]{}%
  }%
  \providecommand\transparent[1]{%
    \errmessage{(Inkscape) Transparency is used (non-zero) for the text in Inkscape, but the package 'transparent.sty' is not loaded}%
    \renewcommand\transparent[1]{}%
  }%
  \providecommand\rotatebox[2]{#2}%
  \newcommand*\fsize{\dimexpr\f@size pt\relax}%
  \newcommand*\lineheight[1]{\fontsize{\fsize}{#1\fsize}\selectfont}%
  \ifx\svgwidth\undefined%
    \setlength{\unitlength}{386.09262085bp}%
    \ifx\svgscale\undefined%
      \relax%
    \else%
      \setlength{\unitlength}{\unitlength * \real{\svgscale}}%
    \fi%
  \else%
    \setlength{\unitlength}{\svgwidth}%
  \fi%
  \global\let\svgwidth\undefined%
  \global\let\svgscale\undefined%
  \makeatother%
  \begin{picture}(1,0.60430874)%
    \lineheight{1}%
    \setlength\tabcolsep{0pt}%
    \put(0,0){\includegraphics[width=\unitlength,page=1]{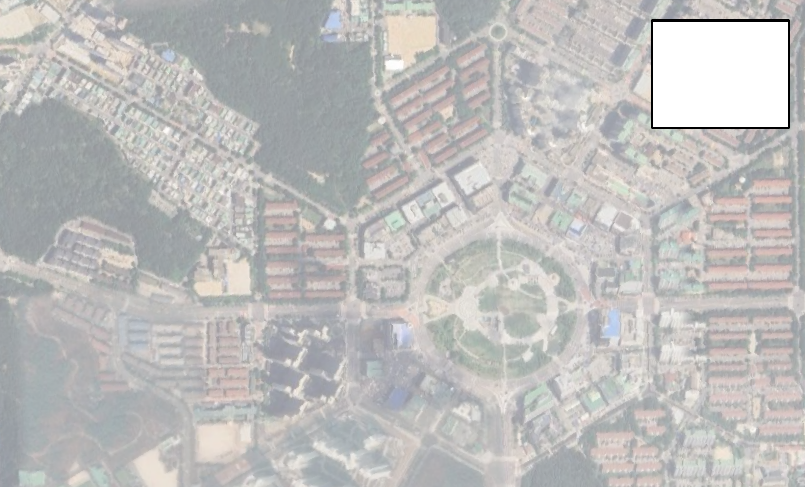}}%
    \put(0.85921939,0.45625195){\makebox(0,0)[lt]{\lineheight{1.25}\smash{\begin{tabular}[t]{l}Ouster\end{tabular}}}}%
    \put(0.81800612,0.55129365){\makebox(0,0)[lt]{\lineheight{1.25}\smash{\begin{tabular}[t]{l}$>$ 2700 scans\\$>$ 9\,km Length\\\end{tabular}}}}%
    \put(0,0){\includegraphics[width=\unitlength,page=2]{motivation.pdf}}%
    \put(0.0407446,0.16949752){\makebox(0,0)[lt]{\lineheight{1.25}\smash{\begin{tabular}[t]{l}Time\end{tabular}}}}%
    \put(0,0){\includegraphics[width=\unitlength,page=3]{motivation.pdf}}%
    \put(0.84634474,0.07149912){\makebox(0,0)[lt]{\lineheight{1.25}\smash{\begin{tabular}[t]{l}Same Set\end{tabular}}}}%
    \put(0.8224047,0.04581887){\makebox(0,0)[lt]{\lineheight{1.25}\smash{\begin{tabular}[t]{l}Of Parameters\end{tabular}}}}%
    \put(0.02498708,0.55862752){\makebox(0,0)[lt]{\lineheight{1.25}\smash{\begin{tabular}[t]{l}Avia\end{tabular}}}}%
    \put(0.02942993,0.09286542){\makebox(0,0)[lt]{\lineheight{1.25}\smash{\begin{tabular}[t]{l}Avia\end{tabular}}}}%
    \put(0.25039684,0.55829692){\makebox(0,0)[lt]{\lineheight{1.25}\smash{\begin{tabular}[t]{l}Ouster\end{tabular}}}}%
    \put(0,0){\includegraphics[width=\unitlength,page=4]{motivation.pdf}}%
    \put(0.30043447,0.04622217){\makebox(0,0)[lt]{\lineheight{1.25}\smash{\begin{tabular}[t]{l}Detected\end{tabular}}}}%
    \put(0.27780265,0.01880977){\makebox(0,0)[lt]{\lineheight{1.25}\smash{\begin{tabular}[t]{l}Loop Closure\end{tabular}}}}%
    \put(0,0){\includegraphics[width=\unitlength,page=5]{motivation.pdf}}%
    \put(0.13517095,0.26447765){\makebox(0,0)[lt]{\lineheight{1.25}\smash{\begin{tabular}[t]{l}Time $t_i$\end{tabular}}}}%
    \put(0,0){\includegraphics[width=\unitlength,page=6]{motivation.pdf}}%
    \put(0.03202925,0.14366052){\makebox(0,0)[lt]{\lineheight{1.25}\smash{\begin{tabular}[t]{l}$t_j\,{\gg}\,t_i$\end{tabular}}}}%
    \put(0,0){\includegraphics[width=\unitlength,page=7]{motivation.pdf}}%
    \put(0.86049964,0.48870487){\makebox(0,0)[lt]{\lineheight{1.25}\smash{\begin{tabular}[t]{l}Avia\end{tabular}}}}%
  \end{picture}%
\endgroup%

				
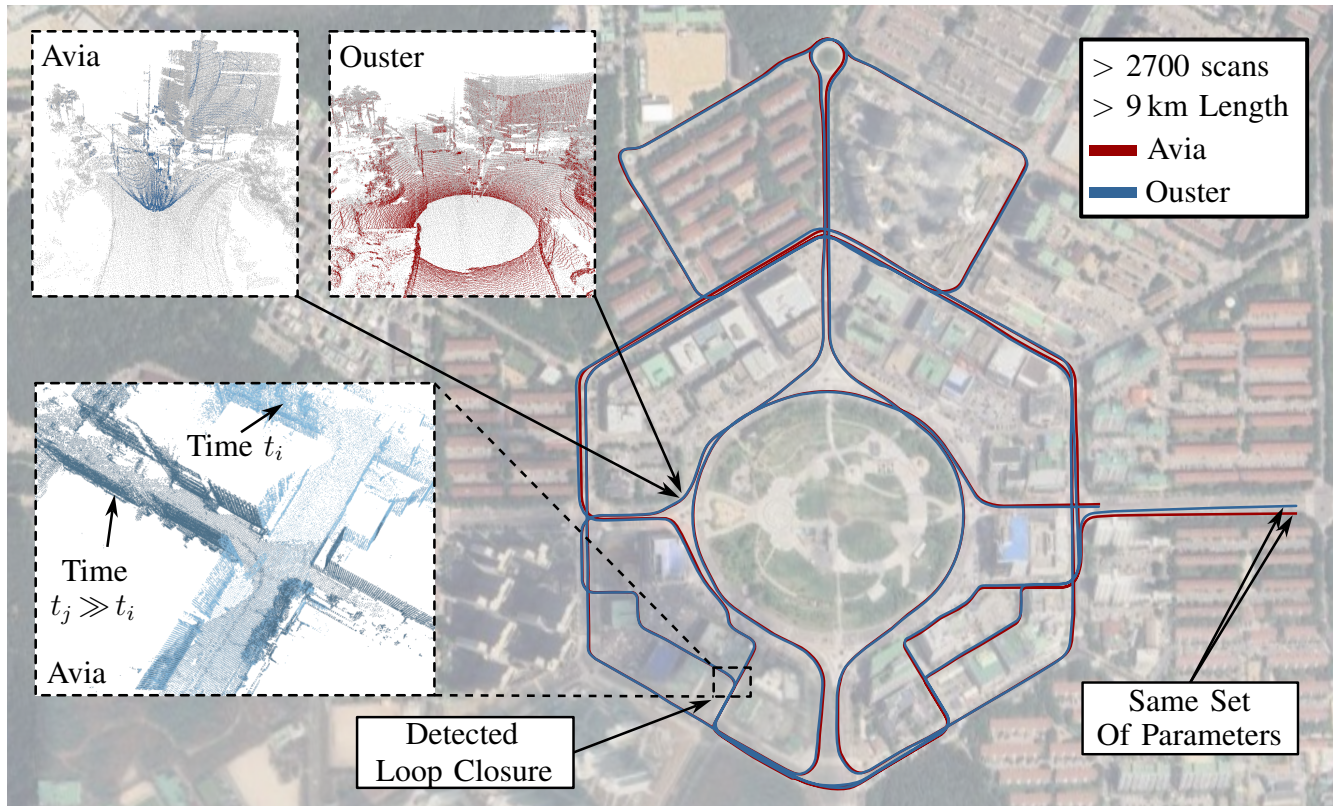
\captionof{figure}{Our proposed SLAM system can accurately estimate the trajectory for a drive of more than 9\,km length in real-time. Using the same parameter configuration, we achieve a similar output for two different LiDAR sensors with different scan patterns and resolutions from the \helipr dataset~\cite{jung2024ijrr}. In both cases, our system can compute the odometry, successfully find loop closures across large time spans, and output a globally consistent trajectory.}
				\label{fig:motivation}
			\end{center}%
		}]
\makeatletter{\renewcommand*{\@makefnmark}{}
\footnotetext{* Authors contributed equally}%
\footnotetext{Tiziano Guadagnino, Benedikt Mersch, Saurabh Gupta, Ignacio Vizzo, and Cyrill Stachniss are with the Center for Robotics, University of Bonn, Germany. Giorgio Grisetti is with Sapienza, University of Rome, Italy. Cyrill Stachniss is additionally with the Department of Engineering Science at the University of Oxford, UK, and with the Lamarr Institute for Machine Learning and Artificial Intelligence, Germany.}%
\footnotetext{This work has partially been funded
	by the Deutsche Forschungsgemeinschaft (DFG, German Research Foundation) under Germany's Excellence Strategy, EXC-2070 -- 390732324 -- PhenoRob,
	by the Deutsche Forschungsgemeinschaft (DFG, German Research Foundation) under STA~1051/5-1 within the FOR 5351~(AID4Crops),
	and
	by the German Federal Ministry of Education and Research (BMBF) in the project ``Robotics Institute Germany'', grant No.~16ME0999.
}

\begin{abstract}
	Robust and accurate localization and mapping of an environment using laser scanners, so-called LiDAR SLAM, is essential to many robotic applications.
	Early 3D LiDAR SLAM methods often exploited additional information from IMU or GNSS sensors to enhance localization accuracy and mitigate drift.
	Later, advanced systems further improved the estimation at the cost of a higher runtime and complexity.
	This paper explores the limits of what can be achieved with a LiDAR-only SLAM approach while following the \kiss (KISS) principle.
	By leveraging this minimalistic design principle, our system, \slam, archives state-of-the-art performances in pose accuracy while requiring little to no parameter tuning for deployment across diverse environments, sensors, and motion profiles.
	We follow best practices in graph-based SLAM and build upon LiDAR odometry to compute the relative motion between scans and construct local maps of the environment.
	To correct drift, we match local maps and optimize the trajectory in a pose graph optimization step.
	The experimental results demonstrate that this design achieves competitive performance while reducing complexity and reliance on additional sensor modalities.
	By prioritizing simplicity, this work provides a new strong baseline for LiDAR-only SLAM and a high-performing starting point for future research.
	Further, our pipeline builds consistent maps that can be used directly for further downstream tasks like navigation.
	Our open-source system operates faster than the sensor frame rate in all presented datasets and is designed for real-world scenarios.
\end{abstract}

\section{Introduction}
\label{sec:intro}
Simultaneous Localization and Mapping (SLAM) is an essential building block for any mobile robot
autonomously navigating in unknown environments~\cite{grisetti2010titsmag, grisetti2020robotics}. Several 3D LiDAR SLAM approaches combine
multiple sensor sources~\cite{shan2020iros, wu2023ral, ruan2020icra}, fusing 3D LiDAR readings with inertial measurement units or GNSS
data~\cite{wu2024icra, xu2021ral-fafr, zheng2022iros, zheng2025tro}. This sensor fusion approach helps reduce tracking errors and enhances the precision of pose
estimation. However, managing multiple sensors in a sensor fusion pipeline can be challenging due to
the need for accurate inter-sensor calibration and time synchronization~\cite{vizzo2023itscws}. Besides that, several recent SLAM
systems make use of neural map representations~\cite{kelly2023icra, pan2024tro, wiesmann2023ral-icra,wiesmann2022ral-iros} or complex architectures~\cite{digiammarino2023ral, digiammarino2022iros,zheng2021ral, dellenbach2022icra}. These
pipelines often overfit to the environment and the specific sensor configuration or motion profile
in use, requiring extensive parameter tuning or training steps to enhance performances in unseen scenarios.

In this paper, we present \slam, a 3D LiDAR-only SLAM pipeline that follows the \kiss principle~\cite{vizzo2023ral, guadagnino2025icra}. The
main goal is to provide a highly performing SLAM system while minimizing the number of components
and parameters. We aim to reduce the system complexity to enhance the generalization of our SLAM system to different environments, sensor resolutions, and motion profiles. Our
approach challenges existing geometric SLAM systems and even modern deep learning-based solutions.
The same system parameters work in various challenging scenarios, such as highway drives of
robot cars, handheld devices, and segways. Furthermore, our method can generalize to different
scanning patterns and scan resolutions.

The main contribution of this paper is a simple yet highly effective approach to LiDAR SLAM that can accurately compute a robot’s pose and the corresponding map online while navigating through an environment.
We identify the core components and adequately evaluate the impact of different modules on such a system.
We show that we obtain highly accurate globally consistent pose estimates while minimizing the number of parameters that require tuning.
In sum, we make three key claims: Our \kiss SLAM approach (i) is on par or better than state-of-the-art SLAM systems in terms of pose accuracy,
(ii) can accurately compute the robot’s pose and map in a large variety of environments, sensor characteristics, and motion profiles with the same system configuration,
and (iii) we can use its mapping output for robot navigation. These claims are backed up by the paper and our experimental evaluation.
We provide an open-source implementation at:~\url{https://github.com/PRBonn/kiss-slam} that precisely follows the description of this paper.

\section{Related Work}
\label{sec:related}
The development of 3D LiDAR-only SLAM emerged to reduce hardware complexity and
enhance applicability in GPS-denied environments~\cite{shan2018iros,deschaud2018icra,
	shan2020iros,jiang2023icra-ccas, wei2022icra} . SuMa~\cite{behley2018rss} is
a pipeline designed for rotating LiDAR scanners based on rendered views from a
surfel map of the environment for data association and loop closure detection.
Other approaches like MULLS~\cite{pan2021icra-mvls} classify points in a scan
based on their geometric features like ground, facades, or pillars and optimize
the pose using different loss functions. Direct SLAM pipelines like MD-SLAM
unify LiDAR and RGB-D sensor processing through dense photometric alignment,
bypassing geometric assumptions and feature
extraction~\cite{digiammarino2022iros}.

CT-ICP~\cite{dellenbach2022icra} demonstrate that LiDAR-centric SLAM can rival
fused systems by refining temporal continuity and motion prediction without
inertial data. It integrates loop closures based on features extracted
on an elevation grid. Although accurate, CT-ICP requires
quite some parameter tuning to operate in unseen environments. Moreover, the
system configuration needs to be adjusted to the specific motion profile of the
platform in use. In contrast, our proposed system does not require parameter tuning
based on the platform motion profile.

Loop closing is an essential functionality for 3D LiDAR SLAM. In our system, we leverage the approach of Gupta~\etalcite{gupta2024icra} for loop closing,
which exploits ORB descriptors~\cite{rublee2011iccv} on a 2D density-based projection of local
maps to re-localize the robot effectively. This approach achieves state-of-the-art performances, requiring little to no parameter tuning for unseen scenarios.

Recent advances in neural representations include PIN-SLAM~\cite{pan2024tro},
which integrates LiDAR odometry with a probability-based implicit neural
mapping framework. This approach achieves accurate 3D reconstruction and loop
closure without pose graph optimization by leveraging signed distance fields
and uncertainty-aware ray casting. These types of systems have a high computational
load and require powerful GPUs to run. As such, currently, these approaches do not operate
at the sensor frame rate when considering the typical hardware setting
available on a real robot. Our system aims to deliver globally accurate pose
estimates while operating at a higher frequency than the sensor frame
rate.

KISS-ICP~\cite{vizzo2023ral} achieves accurate 3D LiDAR odometry through
minimalistic point-to-point ICP, eliminating the need for IMU integration while
achieving real-time performance through adaptive motion compensation. The
system performs impressively with little to no parameter tuning required when
deployed in an unseen environment. Although KISS-ICP performs well in terms of
pose error while tracking~\cite{vizzo2023ral}, the lack of a loop closing and pose graph
optimization module limits its performance. This work aims to extend KISS-ICP
to a complete SLAM pipeline. Following the \kiss principle, we design a system
that requires low parameter tuning, delivers robust and accurate performances,
and has real-time capabilities.
\section{Our Approach to LiDAR-based SLAM}
\label{sec:main}
This section presents the main components of~\slam, our proposed LiDAR-based SLAM system. Given a new LiDAR scan, we first estimate the odometry of the platform based on sensor data in~\secref{sec:odom}.
We then combine the scan measurement and the ego-motion information into our local mapping module in~\secref{sec:mapping}.
Upon completing a map, we check for loop closures with previous local maps in~\secref{sec:closure}.
If a closure is detected, we perform a pose graph optimization of the local map reference frames.
Finally, we output the resulting map as a 3D occupancy grid in~\secref{sec:occupancy}. \figref{fig:overview} provides a visualization of our pipeline.

\begin{figure*}[t]
	\centering
	\def\svgwidth{0.99\linewidth}
	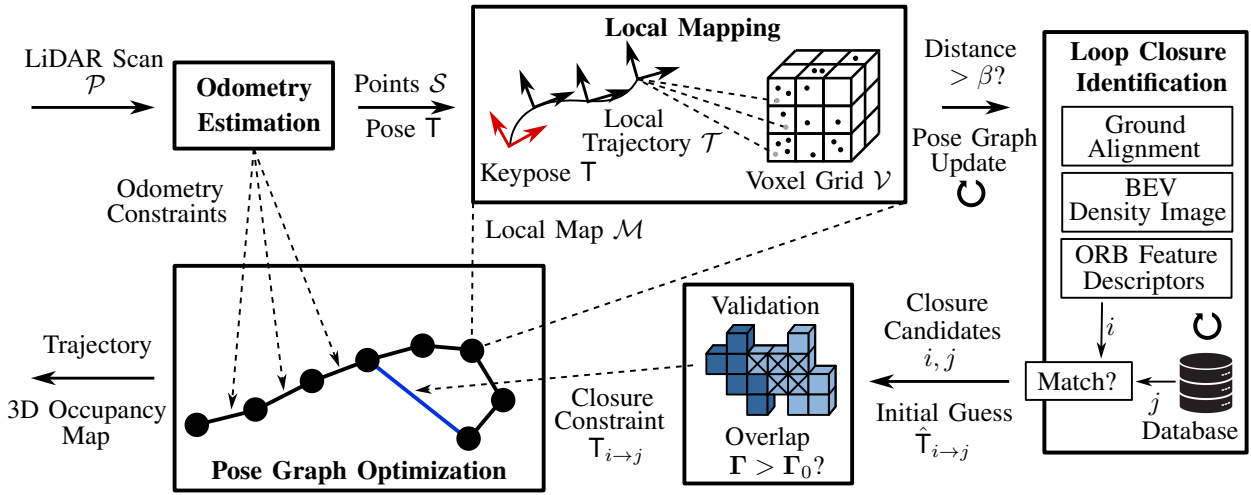
	\caption{Overview of KISS-SLAM. Our pipeline processes each LiDAR scan and first computes the odometry of the scanner, see~\secref{sec:odom}. Successively, the system integrates the LiDAR point cloud with the motion estimate into a local map in~\secref{sec:mapping}. When the LiDAR moton exceeds~$\beta$ meters, the system updates the pose graph and searches for loop closures, see~\secref{sec:closure}. If they are positively validated based on overlap criteria, we perform a pose graph optimization step.}
	\label{fig:overview}
\end{figure*}

\subsection{LiDAR Odometry Using KISS-ICP}\label{sec:odom}
To obtain the pose~$\pose_t\,{\in}\,\se(3)$ of the LiDAR sensor in the odometry frame at time~$t$, we first pre-process the incoming point cloud~$\set{P}\,{=}\,\{ \d{p}_i \,{\mid}\, \d{p}_i \,{\in}\, \RR^3 \}$ expressed in the sensor frame by de-skewing and voxel downsampling resulting in~$\set{S}\,{=}\,\{ \spoint_i \,{\mid}\, \spoint_i \,{\in}\, \RR^3 \}$, see~\cite{vizzo2023ral}. Given the previous estimate of the LiDAR pose~$\pose_{t-1}$ and a constant velocity motion model prediction~$\cv\,{\in}\,\se(3)$, we compute an initial guess for the current LiDAR pose as
\begin{equation}
	\pred_{t} = \pose_{t-1}\,\cv.
\end{equation}

We then refine this estimate by using the point-to-point~\ac*{icp} algorithm. At each iteration, we obtain a set of correspondences between the source~$\set{S}$ and our local map points~$\set{Q} \,{=}\, \{ \d{q}_i \,{\mid}\, \d{q}_i \,{\in}\, \RR^3\}$ which are stored in a voxel grid as in KISS-ICP~\cite{vizzo2023ral} and updated after registration. We define the residual~$\v{r}$ between the point~$\tpoint$ and the point~$\spoint$ transformed by~$\pose$ as
\begin{align}
	\label{eq:residual}
	\v{r}(\pose) = \pose \spoint - \tpoint.
\end{align}

We then define our point-to-point cost function as:
\begin{align}
	\label{eq:cost}
	\rchi(\pred_{t}) = \sum_{\left(\spoint,\tpoint\right) \in \set{C}} \left\| \v{r}\left(\pred_{t}\right) \right\|^{2}_{2},
\end{align}
where~$\set{C}$ is the set of nearest neighbor correspondences. We can then minimize~\eqref{eq:cost} in a least squares fashion as:
\begin{align}
	\label{eq:optimization}
	\pert = \argmin_{\pert} \rchi(\pred_{t}\boxplus\pert),
\end{align}
where~$\pert\,{\in}\,\RR^6$ is the correction vector, and~$\boxplus$ applies the correction vector to the current pose estimate. We repeat this process, including nearest neighbor correspondence search and least squares optimization, until convergence, resulting in the new pose estimate~$\pose_{t}$. After convergence, we update the local map with a downsampled version of the registered scan.

\subsection{Local Mapping and Pose Graph Construction}\label{sec:mapping}
Like several existing methods, our approach avoids maintaining a single global map through a local map-splitting strategy. This design exploits the inherent local consistency of our odometry estimates, which yields precise short-term trajectories within individual map segments. In our system, each local map~$\localmap_k$ contains a keypose reference frame~$\keypose_k\,{\in}\,\se(3)$, an odometry-derived local trajectory~$\trajectory_k\,{\in}\,\{\pose_{t\rightarrow t+1} \,{\mid}\, t_s\leq t < t_e \}$, and a keypose-centered voxel grid~$\grid_k$ containing the local map points. Here,~$t_s$ and~$t_e$ denote the beginning and end times of the local map's construction.

We first integrate subsampled, deskewed scans into the current local map using the odometry motion estimation presented in~\secref{sec:odom}. Next, we evaluate the total distance from the sensor's estimated pose with respect to the keypose~$\keypose_k$ of the current local map~$\localmap_k$. We create a new local map when the traveled distance exceeds a certain threshold~$\beta$. The keypose of this new local map is initialized with the current global pose estimate which is a combination of the previous keypose and the current odometry estimate. Then, we reset the odometry frame and initialize the new voxel grid with a spatially cropped version of the previous local map.

The system maintains a pose graph where nodes represent keypose-anchored local maps connected through odometric edge constraints. Subsequent sections detail our loop closure detection methodology and explain how we integrate them into the pose graph to enforce global consistency.
\subsection{Loop Closing}\label{sec:closure}
After splitting the local maps, we search for loop closures between our last local map and all previous local maps generated. Such loop closures provide additional constraints for the pose graph to correct the drift from odometry toward global consistency.

To search for a closure, we use the approach by Gupta~\etalcite{gupta2024icra,gupta2025arxiv}. First, we identify the ground points in the local map and align the point cloud with the xy-plane of the keypose reference frame. Next, we project the local map into a bird's eye view representation by computing the density of projected 3D LiDAR points for each 2D grid cell. We compute binary ORB~\cite{rublee2011iccv} feature descriptors from the 2D density image and search for matches in a database of descriptors from all previous local map density images. When we find a loop closure candidate, we perform a RANSAC-based geometric validation that provides a 2D alignment of the density images. We combine it with the initial ground alignment to obtain an initial alignment of the matched local maps~$\closurehat$.

To ensure that we can effectively use the detected loop closure to correct for odometry drift, we perform a validation step based on the 3D information stored in the local maps. To this end, for each voxel in a local map, we compute the mean~$\v{\mu}_i$ of the point coordinates and a per-voxel normal vector~$\v{n}_i$ based on a principal component analysis within the voxel points, resulting in a point cloud~$\set{N}_{k}\,{=}\,\{ \{\v{\mu}_i, \mathbf{n}_i\} \,{\mid}\, \v{\mu}_{i}, \mathbf{n}_i \,{\in}\, \RR^3\}$. We apply the initial guess~$\closurehat$ and perform a point cloud registration step between the voxel-based point clouds~$\set{N}_{i}$ and~$\set{N}_{j}$ that correspond to the local maps involved in the closure, resulting in a transformation~$\closure$. We then compute the Szymkiewicz-Simpson overlap coefficient~\cite{vijaymeena2016mlaij} between the point cloud as:
\begin{equation}
	\label{eq:overlap}
	\boldsymbol\Gamma(\set{N}_{i}, \set{N}_{j}, \closure ) = \frac{|\set{N}_{j} \cap \closure \oplus \set{N}_{i}|}{\min(|\set{N}_{j}|,|\set{N}_{i}|)},
\end{equation}
where~$\pose \oplus \set{N}$ applies the transformation~$\pose$ to the point cloud~$\set{N}$,~$\cap$ indicates the intersection between the point clouds based on the voxel size, and~$|\set{N}|$ is the size of the point cloud. We accept a loop closure in the optimization if the value of~$\boldsymbol\Gamma$ is above a certain threshold~$\boldsymbol\Gamma_{0}$, which we set fixed to~40\,\%. In that case, we add a loop closure edge based on~$\closure$ and optimize the pose graph to obtain the most up-to-date estimate of the keyposes.

\subsection{Fine Grained Pose Graph Optimization}\label{sec:occupancy}
After having processed all scans, we perform a fine-grained pose graph optimization by fixing the nodes corresponding to the keyposes and add new nodes and edges using the local trajectories stored in the local maps.
This way, we can optimize the scan poses within the local maps, allowing us to redistribute minor drift errors among the poses in a local chunk of trajectory. This procedure can be seen as an offline implementation of the approach of Grisetti~\etalcite{grisetti2010icra}.

\section{Experimental Evaluation}
\label{sec:exp}
The main focus of this paper is a simple yet highly effective approach for LiDAR SLAM that can accurately estimate a robot’s pose and the corresponding map of the environment while the platform navigates through it. The experiments reported here support our key claims, which are that our \slam approach is (i) on par or better than state-of-the-art SLAM systems in terms of pose accuracy, (ii) can accurately compute the robot’s pose and map in a large variety of environments, sensor characteristics, and motion profiles with the same system configuration, and (iii) we demonstrate how to use the mapping output of the system for a robotics navigation task.

\subsection{Experimental Setup}
We employ various datasets and established evaluation methodologies to assess our system's performance. To examine our approach's effectiveness across different autonomous driving datasets utilizing diverse sensors, we evaluate it on the MulRan~\cite{kim2020icra}, \helipr~\cite{jung2024ijrr}, and Apollo~\cite{huang2018cvprws} datasets. \helipr, in particular, presents four different LiDAR sensors with different ranging technologies and scanning patterns. We exclude \helipr sequences recorded with the Velodyne VLP-16 because of its self-occlusion with surrounding sensors. Furthermore, we demonstrate our method's versatility by applying it to a highly dynamic motion profile, such as those presented in the NCLT~\cite{carlevaris-bianco2016ijrr} dataset, which uses a Segway platform, and the Newer College dataset~\cite{ramezani2020iros}, captured with a handheld device.

\begin{table}[t]
	\centering
	\begin{tabular}{lc}
		\toprule
		Method   & \makecell{2012-01-08}           \\
		\midrule
		PIN-SLAM & \textbf{1.69}\,/\,\textbf{0.30} \\
		SuMa     & 316.48\,/\,50.41                \\
		CT-ICP   & -\,/\,-                         \\
		MULLS    & 6.44\,/\,\underline{0.59}       \\
		Ours     & \underline{3.00}\,/\,0.61       \\
		\bottomrule
	\end{tabular}
	\caption{Quantitative results on the NCLT dataset, we report the~\ac*{ate} in meters and the relative KITTI odometry metric in percentage as [m] / [\%]. The best and second best performing methods are reported in \textbf{bold} and \underline{underline}, respectively.}
	\label{tab:NCLT}
\end{table}

\begin{table}[t]
	\centering
	\begin{tabular}{lcccc}
		\toprule
		Method   & \makecell{DCC}                  & \makecell{KAIST}                & \makecell{Riv}                     & \makecell{Sej}                    \\
		\midrule
		PIN-SLAM & \textbf{3.57}\,/\,\textbf{1.37} & \textbf{2.42}\,/\,\textbf{0.32} & \textbf{7.55}\,/\,\underline{0.46} & \underline{782.96}\,/\,7.68       \\
		SuMa     & 39.41\,/\,4.15                  & 26.65\,/\,1.34                  & -\,/\,-                            & -\,/\,-                           \\
		CT-ICP   & 4.03\,/\,\underline{1.43}       & \underline{2.76}\,/\,0.50       & 8.37\,/\,0.58                      & -\,/\,-                           \\
		MULLS    & 27.30\,/\,1.72                  & 34.21\,/\,0.60                  & 66.60\,/\,2.27                     & 1082.55\,/\,\underline{2.53}      \\
		Ours     & \underline{3.67}\,/\,1.44       & 2.98\,/\,\underline{0.34}       & \underline{7.96}\,/\,\textbf{0.40} & \textbf{178.88}\,/\,\textbf{0.25} \\
		\bottomrule
	\end{tabular}
	\caption{Quantitative results on the MulRan dataset, we report the~\ac*{ate} in meters and the relative KITTI odometry metric in percentage as [m] / [\%]. The best and second best performing methods are reported in \textbf{bold} and \underline{underline}, respectively.}
	\label{tab:AvgMulRan}
\end{table}

\subsection{Pose Accuracy Evaluation}
\begin{table*}[t]
	\centering
	\begin{tabular}{lccccccccc}
		\toprule
		Method   & \makecell{Bri                                                                                                                                                                                                                                                                                                                                              \\ Aeva} & \makecell{Bri \\ Avia} & \makecell{Bri \\ Ouster} & \makecell{Rou \\ Aeva} & \makecell{Rou \\ Avia} & \makecell{Rou \\ Ouster} & \makecell{To \\ Aeva} & \makecell{To \\ Avia} & \makecell{To \\ Ouster} \\
		\midrule
		PIN-SLAM & -\,/\,-                          & 365.72\,/\,2.15                         & \textbf{19.23}\,/\,\textbf{0.13}       & -\,/\,-                                & 7.02\,/\,0.51                      & \underline{1.47}\,/\,\textbf{0.11} & 41.19\,/\,\underline{1.03} & \textbf{11.40}\,/\,\underline{0.64} & \underline{2.55}\,/\,\textbf{0.12} \\
		SuMa     & -\,/\,-                          & -\,/\,-                                 & 140.01\,/\,1.08                        & -\,/\,-                                & -\,/\,-                            & 14.88\,/\,0.46                     & 132.17\,/\,3.16                        & -\,/\,-                             & -\,/\,-                            \\
		CT-ICP   & -\,/\,-                          & \textbf{41.47}\,/\,\textbf{0.36}        & 579.62\,/\,0.89                        & \underline{10.04}\,/\,\underline{0.72} & \textbf{3.36}\,/\,\underline{0.25} & 1.81\,/\,\underline{0.16}          & 65.29\,/\,1.37                         & 63.72\,/\,2.02                      & -\,/\,-                            \\
MULLS & \underline{356.06}\,/\,\underline{9.17} & 321.87\,/\,4.06 & 52.65\,/\,0.68 & 19.08\,/\,1.43 & 16.39\,/\,1.06 & 2.65\,/\,0.24 & \underline{39.82}\,/\,2.85 & 14.93\,/\,1.35 & 4.45\,/\,0.27 \\ 
		Ours     & \textbf{98.61}\,/\,\textbf{1.57} & \underline{148.88}\,/\,\underline{1.82} & \underline{19.47}\,/\,\underline{0.28} & \textbf{6.06}\,/\,\textbf{0.47}        & \underline{3.84}\,/\,\textbf{0.23} & \textbf{1.18}\,/\,0.17             & \textbf{14.44}\,/\,\textbf{0.85}       & \underline{12.01}\,/\,\textbf{0.46} & \textbf{1.99}\,/\,\underline{0.21} \\
		\bottomrule
	\end{tabular}
	\caption{Quantitative results on the Bridge (Bri), Roundabout (Rou) and Town (To) sequences of the \helipr dataset, we report the~\ac*{ate} in meters and the relative KITTI odometry metric in percentage as [m] / [\%]. The best and second best performing methods are reported in \textbf{bold} and \underline{underline}, respectively.}
	\label{tab:AvgHeLiPR}
\end{table*}

\begin{table*}[t]
	\centering
	\begin{tabular}{lccccccc}
		\toprule
		Method   & \makecell{BTS                                                                                                                                                                                                                                                        \\ 2018-10-12} & \makecell{CP \\ 2018-10-11} & \makecell{H237 \\ 2018-10-12} & \makecell{MAVE \\ 2018-10-12} & \makecell{SB \\ 2018-10-03} & \makecell{SJD \\ 2018-10-11 \\ 1} & \makecell{SJD \\ 2018-10-11 \\ 2} \\
		\midrule
		PIN-SLAM & \underline{6.10}\,/\,\underline{0.24} & \textbf{0.64}\,/\,\textbf{0.16}       & \underline{97.11}\,/\,1.66          & \textbf{7.06}\,/\,\underline{0.20} & \textbf{2.23}\,/\,\textbf{0.25}    & \underline{1.86}\,/\,0.22       & 1.15\,/\,\underline{0.18}          \\
		SuMa     & 181.19\,/\,3.55                       & -\,/\,-                               & 366.66\,/\,9.51                     & 79.68\,/\,0.43                     & -\,/\,-                            & -\,/\,-                         & -\,/\,-                            \\
		CT-ICP   & 10.81\,/\,0.27                        & 0.97\,/\,0.30                         & 258.87\,/\,0.62                     & 61.39\,/\,0.33                     & 667.28\,/\,\underline{1.60}        & \textbf{0.78}\,/\,\textbf{0.16} & \underline{1.09}\,/\,\textbf{0.17} \\
		MULLS    & 104.14\,/\,0.27                       & 47.03\,/\,0.90                        & 354.21\,/\,\textbf{0.39}            & 182.59\,/\,\textbf{0.16}           & -\,/\,-                            & 13.38\,/\,0.32                  & 11.41\,/\,0.56                     \\
		Ours     & \textbf{3.74}\,/\,\textbf{0.23}       & \underline{0.80}\,/\,\underline{0.22} & \textbf{30.16}\,/\,\underline{0.40} & \underline{9.08}\,/\,0.26          & \underline{2.51}\,/\,\textbf{0.25} & 2.00\,/\,\underline{0.17}       & \textbf{0.66}\,/\,\textbf{0.17}    \\
		\bottomrule
	\end{tabular}
	\caption{Quantitative results on the Apollo dataset, we report the~\ac*{ate} in meters and the relative KITTI odometry metric in percentage as [m] / [\%]. The best and second best performing methods are reported in \textbf{bold} and \underline{underline}, respectively.}
	\label{tab:Apollo}
\end{table*}

\begin{table*}[t]
	\centering
	\begin{tabular}{lccccccc}
		\toprule
		Method   & \makecell{2020                                                                                                                                                                                                                              \\ 01-short} & \makecell{2020 \\ 02-long} & \makecell{2021 \\ cloister} & \makecell{2021 \\ math e} & \makecell{2021 \\ quad e} & \makecell{2021 \\ stairs} & \makecell{2021 \\ underground e} \\
		\midrule
		PIN-SLAM & \underline{0.42}\,/\,0.33          & \textbf{0.31}\,/\,\textbf{0.22}       & \underline{0.15}\,/\,0.62       & \textbf{0.08}\,/\,0.29                & \textbf{0.09}\,/\,\underline{0.12} & \textbf{0.06}\,/\,-    & \textbf{0.07}\,/\,-    \\
		SuMa     & 2.06\,/\,9.11                      & 5.77\,/\,3.04                         & 0.17\,/\,0.79                   & 0.16\,/\,0.32                         & 0.21\,/\,0.66                      & 1.85\,/\,-             & \underline{0.11}\,/\,- \\
		CT-ICP   & 0.63\,/\,0.42                      & 25.06\,/\,3.01                        & 0.17\,/\,\underline{0.31}       & \underline{0.09}\,/\,\underline{0.14} & 0.19\,/\,0.14                      & -\,/\,-                & 0.15\,/\,-             \\
		MULLS    & 0.47\,/\,\textbf{0.17}             & 8.47\,/\,4633.62                      & \textbf{0.13}\,/\,\textbf{0.17} & 0.13\,/\,\textbf{0.02}                & \underline{0.16}\,/\,0.34          & \underline{1.82}\,/\,- & 0.69\,/\,-             \\
		Ours     & \textbf{0.30}\,/\,\underline{0.26} & \underline{1.58}\,/\,\underline{2.14} & 0.40\,/\,1.18                   & 0.15\,/\,0.61                         & \underline{0.16}\,/\,\textbf{0.03} & 3.58\,/\,-             & 0.12\,/\,-             \\
		\bottomrule
	\end{tabular}
	\caption{Quantitative results on the Newer College dataset, we report the~\ac*{ate} in meters and the relative KITTI odometry metric in percentage as [m] / [\%]. The best and second best performing methods are reported in \textbf{bold} and \underline{underline}, respectively.}
	\label{tab:NCD}
\end{table*}

\label{sec:performance}
The first experiment evaluates the performance of our approach. It also supports our first claim, namely that our system reports state-of-the-art performance in pose accuracy. We use the standard~\ac*{ate} to measure absolute pose estimation performances. We use the well-known KITTI metric~\cite{geiger2013ijrr} to assess the relative error. We use the open-source implementation from the \textit{evo} package~\cite{grupp2017github} instead of a custom implementation for both metrics. We do this to standardize the results' experimental evaluation and simplify reproducibility. For the MulRan and \helipr datasets, we report the average value of the metrics over each scene, as for both datasets, three different runs are performed per scene.

To assess the performance of our proposed SLAM system, we compare the results to various state-of-the-art open-source LiDAR-only SLAM systems such as SuMa~\cite{behley2018rss}, MULLS~\cite{pan2021icra-mvls}, CT-ICP~\cite{dellenbach2022icra}, and PIN-SLAM~\cite{pan2024tro}. We report the results in~\tabref{tab:NCLT} to~\tabref{tab:NCD}. The results show that our system consistently delivers state-of-the-art performance in terms of pose accuracy, regardless of the motion profile, scanning pattern, and sensor resolution, often ranking first or second. Furthermore, \slam is the only approach that can effectively compute the poses in all the reported scenarios. For example, in~\tabref{tab:AvgHeLiPR}, we can see how our SLAM pipeline can successfully run on the challenging Aeva sensor in \helipr, showcasing the robustness and versatility of our approach, even compared with more sophisticated neural SLAM techniques like PIN-SLAM\@. Notice that we obtain all our results with the same system configuration, and no parameter tuning is applied. Furthermore, KISS-SLAM can compute the pose and map estimate above the sensor frame rate. In contrast, the closest baseline regarding pose accuracy, PIN-SLAM, needs to balance runtime and performance. As investigated by Pan~\etalcite{pan2024tro}, the PIN-SLAM configuration needed to achieve the level of pose accuracy that we report in our analysis does not allow the pipeline to run at sensor frame rate on a single NVIDIA A4000 GPU\@.

\subsection{Ablation on Parameter Change}\label{sec:parameter}
This experiment supports our second claim that our system can accurately estimate the robot's pose and map in many environments, sensor characteristics, and motion profiles with the same configuration. In contrast, existing SLAM approaches usually require fine-tuning parameters to work successfully with different setups.

Note that this section does not evaluate the total number of parameters because this quantity is not straightforward to measure. In fact, the number of parameters exposed in a configuration file might not contain other parameters, some of which are set as constants in the implementation.

In this experiment, we count the parameters that had to be changed between different runs to achieve the results reported in~\tabref{tab:AvgHeLiPR} and~\tabref{tab:NCD}, because they impose challenging changes in the data. For \helipr, we can compare multiple sensor resolutions and scanning patterns using the Avia, Aeva, and Ouster data recorded with the same motion profile in the same environment. We include data recorded with a substantially different motion profile with the Newer College handheld dataset. Note that we also consider data-dependent parameters because, typically, these values influence other parameters in the pipeline~\cite{vizzo2023ral,pan2024tro}.

We can see in~\tabref{tab:parameters} that multiple parameters must be modified for all baselines when running the approaches on the different sensors of the \helipr dataset. We also evaluate the number of parameters that must be changed from running the approach on data acquired by an autonomous vehicle to a dataset recorded with a handheld device. Again, all baselines require a substantial change in the configuration, whereas our approach works out of the box. This experiment illustrates the robust generalization capabilities of our SLAM method, which maintains high performance across diverse sensor configurations, motion profiles, and environments without requiring manual parameter tuning.
\begin{table}[t]
	\centering
	\begin{tabular}{lcccc}
		\toprule
		         & \helipr Ae     & \helipr Ae     & \helipr Av     & \helipr O      \\
		         & $\updownarrow$ & $\updownarrow$ & $\updownarrow$ & $\updownarrow$ \\
		         & \helipr Av     & \helipr O      & \helipr O      & NCD-2020       \\
		\midrule
		PIN-SLAM & 12             & 11             & 13             & 23             \\
		SuMa     & 7              & 9              & 9              & 19             \\
		CT-ICP   & 16             & 16             & 16             & 16             \\
		MULLS    & -              & -              & \textbf{0}     & 25             \\
		Ours     & \textbf{0}     & \textbf{0}     & \textbf{0}     & \textbf{0}     \\
		\bottomrule
	\end{tabular}
	\caption{Number of changed parameters between configurations for different sensors and motion profiles. For \helipr, ``Ae'' denotes the Aeva sensor, ``Av'' is Avia, and ``O'' is the Ouster scanner. Note that ``-'' means that there is no working set of parameters for at least one of the sequences. The lower the number of changed parameters, the easier it is to run the approach on new data. The lowest number of changed parameters is in \textbf{bold}.}
	\label{tab:parameters}
\end{table}

\subsection{Navigation Experiment on Real Robot}
Finally, our last experiment supports our third claim that we can use the output of~\slam to perform robotic navigation tasks. In essence, we want to illustrate the usability of a global map computed using the pose estimates from our approach. To this end, we build a 2D occupancy grid map of an office-like environment and perform a global localization experiment. This demonstrates the general applicability of our approach when using different platforms for mapping and localization.

We start by recording a sequence of 3D LiDAR scans using a Hesai XT-32 scanner mounted on a Clearpath Husky robot. We run~\slam on this sequence and generate a 3D occupancy grid map of the environment, with a resolution of~$0.05$\,m per voxel, following a standard occupancy grid mapping algorithm~\cite{thrun2005probrobbook}.
For this experiment, we reduce the max range of the points that~\slam processes from $100$ meters to $50$ meters, as the office environment is relatively small compared to an outdoor scene.
Other than this value, we leave the pipeline configuration untouched.

We would like to point out that \slam could run faster than the sensor frame rate on the robot's onboard computer, an Intel NUC equipped with an Intel i7 processor and 32\,GB of RAM. Additionally, we run the grid mapping pipeline on the same computer.

To create a 2D occupancy grid, we process a horizontal slice of the 3D occupancy map within a specific height range,~$[z_{min},z_{max}]$. We assign the 2D cells occupancy values based on the maximum occupancy within the corresponding vertical column of the horizontal slice. Once we create a map this way, we aim to globally localize and track the pose of a second robot, the Clearpath Dingo, equipped with a SICK TiM781S 2D LiDAR. Since the 2D LiDAR is mounted at~$0.16$\,m above the ground, we initially set~$z_{min}=0.1$\,m and~$z_{max}=0.2$\,m for the 2D occupancy grid construction. The resulting maps are depicted in~\figref{fig:map}. The final 2D map is shown in~\figref{fig:map}.
We use a \ac*{mcl} implementation from the \ac*{srrg}~\cite{grisetti2018github}, which we abbreviate here with \ac*{srrg}-Loc, to perform localization with the Dingo equipped with the 2D LiDAR. To measure the quality of our occupancy map, we compare the localization performances of \ac*{srrg}-Loc obtained using our map against an occupancy grid generated with GMapping~\cite{grisetti2007tro}. This second map is based on data recorded in the same environment but with the Dingo robot and its 2D LiDAR.

For our evaluation, we assess global localization performance and pose-tracking accuracy after convergence. To achieve this, we initialize \ac*{srrg}-Loc with 10,000 particles uniformly distributed across the free space in the occupancy map. We define the convergence criterion based on the localizer’s reported pose covariance. Specifically, we require the standard deviation of the largest principal component of the translational part to be below 0.2\,m and the standard deviation of the heading angle to be under 10°. We measure the convergence time that is required for the estimate to converge. Once convergence is achieved, we evaluate the pose tracking accuracy using ground truth poses obtained by detecting AprilTags~\cite{olson2011icra-aara} placed on the office's ceiling with an upwards-looking camera. To ensure robustness, we run \ac*{srrg}-Loc 10 times on each sequence, each time with a randomly initialized seed. We provide the results, including the mean and standard deviation over these 10 runs for each sequence in \tabref{tab:localization}. The results show no significant difference between the two map representations, showcasing that our system can effectively generate a map suitable for global localization and tracking, validating the real-world applicability of~\slam.

\begin{table*}[ht]
	\centering
	\begin{tabular}{cccccccc}
		\toprule
		                                    &             & \multicolumn{2}{c}{Global Localization} & \multicolumn{4}{c}{Pose Tracking: Absolute Trajectory Error (ATE)}                                                                                                                                                         \\
		Sequence                            & SLAM Method & Convergence                             & Success                                                            & \multicolumn{2}{c}{ATE translation\,[cm]} & \multicolumn{2}{c}{ATE rotation\,[°]}                                                                     \\
		                                    &             & Time\,[s] $\downarrow$                  & Rate\,[\%] $\uparrow$                                              & RMS $\downarrow$                          & Max $\downarrow$                      & RMS $\downarrow$                & Max $\downarrow$                \\
		\midrule
		\multirow{2}{*}{Static Sequence 1}  & GMapping    & $11.27\pm5.68$                          & 100                                                                & $\textbf{9.48}\pm0.05$                    & $16.97\pm\textbf{0.11}$               & $\textbf{1.89}\pm\textbf{0.01}$ & $\textbf{6.49}\pm\textbf{0.07}$ \\
		                                    & Ours        & $\textbf{10.34}\pm\textbf{4.14}$        & 100                                                                & $9.71\pm\textbf{0.02}$                    & $\textbf{16.01}\pm0.13$               & $1.97\pm\textbf{0.01}$          & $7.81\pm\textbf{0.07}$          \\
		\midrule
		\multirow{2}{*}{Static Sequence 2}  & GMapping    & $\textbf{5.68}\pm1.37$                  & 100                                                                & $9.88\pm\textbf{0.02}$                    & $\textbf{15.70}\pm\textbf{0.20}$      & $\textbf{1.77}\pm\textbf{0.01}$ & $\textbf{5.33}\pm\textbf{0.06}$ \\
		                                    & Ours        & $6.19\pm\textbf{0.88}$                  & 100                                                                & $\textbf{9.26}\pm0.05$                    & $17.85\pm2.45$                        & $1.90\pm\textbf{0.01}$          & $7.61\pm\textbf{0.06}$          \\
		\midrule
		\multirow{2}{*}{Static Sequence 3}  & GMapping    & $\textbf{10.51}\pm1.93$                 & 100                                                                & $9.68\pm\textbf{0.13}$                    & $16.42\pm\textbf{0.11}$               & $1.80\pm\textbf{0.01}$          & $\textbf{4.99}\pm\textbf{0.07}$ \\
		                                    & Ours        & $11.73\pm\textbf{1.73}$                 & 100                                                                & $\textbf{8.51}\pm0.20$                    & $\textbf{14.23}\pm0.43$               & $\textbf{1.79}\pm0.06$          & $5.55\pm0.66$                   \\
		\midrule
		\multirow{2}{*}{Dynamic Sequence 1} & GMapping    & $11.81\pm5.05$                          & 90                                                                 & $\textbf{5.71}\pm0.18$                    & $\textbf{11.06}\pm0.37$               & $\textbf{1.81}\pm\textbf{0.01}$ & $\textbf{6.74}\pm0.09$          \\
		                                    & Ours        & $\textbf{11.25}\pm\textbf{1.87}$        & 90                                                                 & $9.57\pm\textbf{0.12}$                    & $17.54\pm\textbf{0.08}$               & $1.99\pm\textbf{0.01}$          & $6.80\pm\textbf{0.07}$          \\
		\midrule
		\multirow{2}{*}{Dynamic Sequence 2} & GMapping    & $\textbf{14.06}\pm\textbf{5.52}$        & 100                                                                & $\textbf{7.43}\pm0.10$                    & $\textbf{17.60}\pm0.36$               & $\textbf{2.38}\pm\textbf{0.03}$ & $7.99\pm\textbf{0.08}$          \\
		                                    & Ours        & $22.66\pm7.34$                          & 100                                                                & $10.42\pm\textbf{0.05}$                   & $23.73\pm\textbf{0.17}$               & $2.67\pm0.05$                   & $\textbf{7.92}\pm0.11$          \\
		\bottomrule
	\end{tabular}
	\caption{Quantitative evaluation of 2D localization performance using our SLAM-generated 2D occupancy grid map and comparison with a GMapping grid map. The evaluation is based on five sequences recorded in an office environment containing static and dynamic scenes. We report localization metrics' mean and standard deviation over 10 runs of \ac*{srrg}-Loc. The best results are in \textbf{bold}.}
	\label{tab:localization}
\end{table*}

\begin{figure*}[t]
	\centering
	\fontsize{8}{8}\selectfont
	\def\svgwidth{0.85\linewidth}
	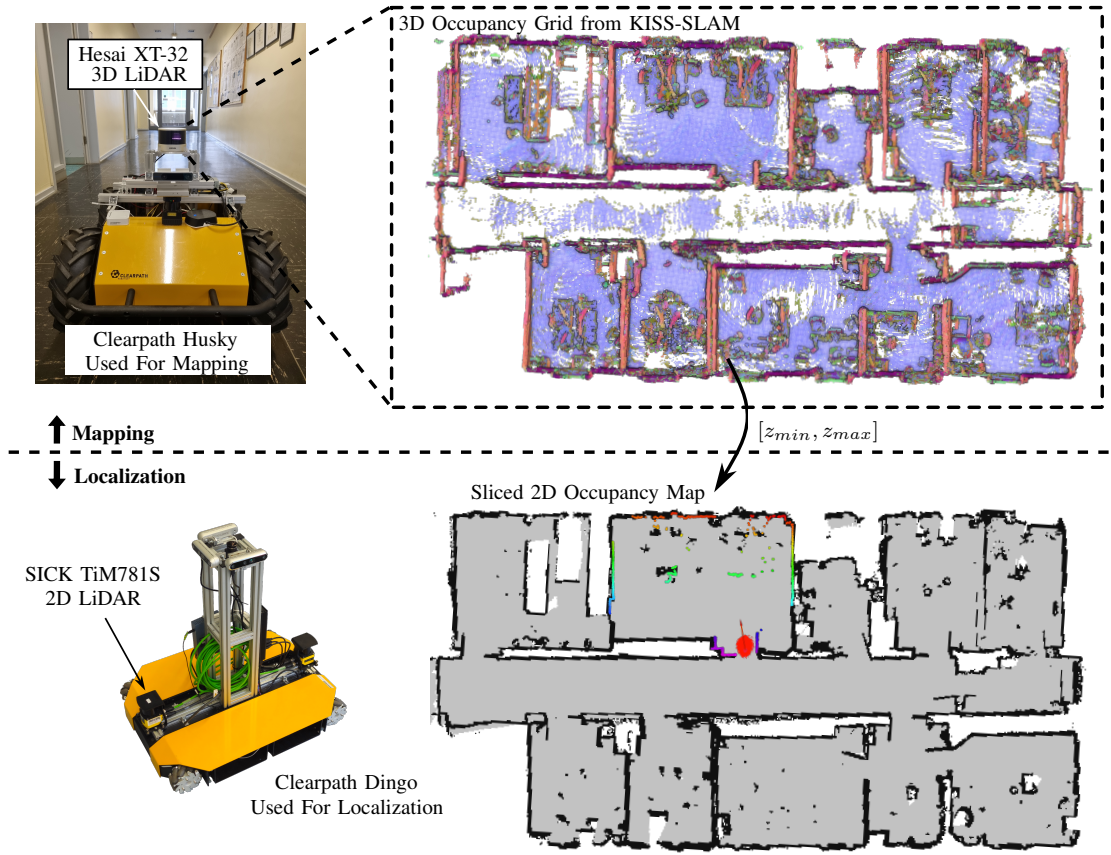
	\caption{Occupancy grid maps generated with our approach. The top shows the 3D occupancy grid we generate using~\slam, after processing all the 3D scans from an Hesai XT-32 with a Clearpath Husky. The bottom displays the corresponding 2D grid generated by slicing the 3D map. We then use this map to do 2D localization using the approach of Grisetti~\etalcite{grisetti2018github}.}
	\label{fig:map}
	\vspace{-0.2cm}
\end{figure*}

\section{Conclusion}
\label{sec:conclusion}
This paper presents \slam, a simple yet highly effective approach to LiDAR SLAM. Our approach
operates solely on LiDAR scans and does not require additional sensors to compute the robot's
trajectory and map of the environment. Our approach exploits a minimalistic design that can be
employed in different challenging environments, such as highway drives, handheld devices, and segways.
Moreover, the system is not tailored to specific range-sensing technologies or scanning
patterns. We only assume that point clouds are generated sequentially as the robot moves through the
environment. We implemented and evaluated our approach on different datasets, provided comparisons
to other existing techniques, supported all claims made in this paper, and released our code. The
experiments suggest that our approach is on par or better than substantially more sophisticated
state-of-the-art LiDAR SLAM systems but relies only on a few parameters and performs well on various
datasets under different conditions with the same parameter set. Furthermore, the system's output can be effectively used to perform downstream tasks like robot navigation. Finally, our
system operates faster than the sensor frame rate in all presented datasets. We believe this work
will be a new baseline for future LiDAR SLAM systems and a high-performing starting point for future
approaches. Our open-source code is robust and simple, easy to extend, and performs well, pushing
the state-of-the-art LiDAR SLAM to its limits and challenging more sophisticated systems.

\section*{Acknowledgments}
We thank Luca Lobefaro, Meher Malladi, and Niklas Trekel for fruitful discussions and Jens Behley and Yue Pan for providing the baseline results.
\clearpage

\bibliographystyle{plain_abbrv}
\bibliography{glorified,new}

\IfFileExists{./certificate/certificate.tex}{
	\subfile{./certificate/certificate.tex}
}{}
\end{document}